\documentclass[a4paper,conference]{IEEEtran}
\IEEEoverridecommandlockouts
\usepackage{cite}
\usepackage{amsmath,amssymb,amsfonts}
\usepackage{graphicx}
\usepackage{textcomp}
\usepackage{xcolor}
\usepackage{tcolorbox}
\usepackage{hyperref}
\usepackage{url}
\usepackage{enumitem}
\usepackage{booktabs}
\usepackage{multirow}
\usepackage{multicol}
\usepackage{helvet}
\def\BibTeX{{\rm B\kern-.05em{\sc i\kern-.025em b}\kern-.08em
    T\kern-.1667em\lower.7ex\hbox{E}\kern-.125emX}}

\begin{document}

\title{Entity-Aware Evaluation of OCR Systems for Historical Document Transcription
}
\author{
\IEEEauthorblockN{
Camilo Mariño$^{1, 2}$$^{\dagger}$,
Diego Belzarena$^{1, 2}$$^{\dagger}$,
Marina Gardella$^{2}$$^{\dagger}$,\\
Ignacio Ramírez$^{1}$,
Gregory Randall$^{1}$,
Jean-Michel Morel$^{3}$
}

\IEEEauthorblockA{$^{1}$\textit{Facultad de Ingeniería, Universidad de la República, Montevideo, Uruguay}}

\IEEEauthorblockA{$^{2}$\textit{Université Paris-Saclay, ENS Paris-Saclay, CNRS, Centre Borelli, France}}

\IEEEauthorblockA{$^{3}$\textit{Lingnan University, Hong Kong}}
}

\maketitle

\begingroup
\renewcommand{\thefootnote}{\dag}
\footnotetext{These authors contributed equally.}
\endgroup

\begin{abstract}
Optical Character Recognition (OCR) is essential for digitizing historical archives, yet standard evaluation metrics such as Character Error Rate (CER) do not necessarily reflect whether meaningful information, such as names, dates, and identifiers, is preserved. In this work, we investigate entity-level analysis of OCR errors as a complementary perspective for evaluating OCR quality in historical documents. Using a publicly available benchmark of 175 documents from the Berrutti Archive, augmented with 5,209 manually annotated entity mentions that will be released upon acceptance, we compare sequence-recognition OCR systems, OCR-specialized vision-language models, and general-purpose vision-language models.

Beyond global transcription accuracy, we analyze entity preservation and recoverability under varying character-level edit tolerances. Our experiments show that OCR rankings depend on the evaluation criterion: while Qwen 3.6 achieves the highest entity coverage across all tested tolerances, several systems with lower exact-match coverage recover substantially when small errors are tolerated, suggesting that many of their errors correspond to localized character-level deviations rather than fundamentally incorrect entity transcriptions. Conversely, some systems remain unable to recover many entities even under relaxed matching. These findings show that OCR quality cannot be fully characterized by aggregate transcription metrics alone and highlight the importance of evaluating how OCR errors affect the preservation and recoverability of meaningful information.

\end{abstract}


\section{Introduction}

Optical Character Recognition (OCR) has become fundamental for making large historical document collections accessible. As OCR systems continue to improve and diversify, reliable evaluation has become increasingly important for comparing methods and guiding their adoption in real-world digitization projects. To this end, OCR quality is commonly assessed using edit-distance metrics such as Character Error Rate (CER) and Word Error Rate (WER), which quantify transcription fidelity at the character and word levels.

The appropriate evaluation of an OCR transcription depends on its intended application. Studies in historical linguistics or paleography often require faithful preservation of the entire document, including spelling variation and orthographic phenomena, making transcription fidelity the primary objective. In contrast, applications such as archival search, information retrieval, genealogy, or historical database construction often prioritize preserving specific information units, particularly person names, dates, locations, organizations, and identifiers, which commonly serve as retrieval keys or database records. Recent work has advocated complementing traditional edit-distance metrics with evaluation strategies that better reflect the intended use of OCR transcriptions~\cite{beyondCERWER}.

In particular, several studies have investigated how OCR errors affect downstream applications, including named entity recognition, entity linking, information retrieval, and retrieval-augmented generation, showing that character-level accuracy alone does not fully explain application performance~\cite{assessingOCRNLP,ocrNERLinking,beyondCERWER,sun2026goodocr}. These works, however, typically assess the complete downstream pipeline rather than the OCR transcription itself. As a result, it remains difficult to disentangle transcription quality from downstream robustness or to determine whether the most relevant information has been faithfully preserved.


In this work, we investigate entity-aware evaluation of OCR systems for historical documents. We annotate named entities in a benchmark dataset from the last Uruguayan dictatorship and evaluate a diverse set of traditional OCR engines, specialized OCR vision-language models, and general-purpose vision-language models (VLMs). Complementing standard CER with Entity CER and entity Coverage, we assess both transcription fidelity and the preservation and recoverability of information-bearing entities. Our experiments show that entity errors are not directly reflected by global transcription accuracy, that entity difficulty varies substantially across categories, and that different OCR systems exhibit distinct preservation and recoverability profiles. These results highlight the importance of evaluating OCR quality according to the information requirements of downstream applications.

\section{Related Work}

\noindent\textbf{Challenges in Historical Document Transcription:}
Recent years have witnessed substantial advances in OCR, driven by both specialized end-to-end recognition systems and vision-language models (VLMs). Large-scale benchmark studies report strong performance across a broad range of OCR systems and document types~\cite{caravani2026}, while recent comparisons on historical collections have shown that modern VLMs can achieve excellent transcription accuracy, albeit with error characteristics that differ substantially from those of traditional OCR models~\cite{vesalainen2026}.

Historical collections nevertheless remain a challenging benchmark due to degraded printing, obsolete typefaces, spelling variation, complex layouts, and heterogeneous acquisition conditions. Recent surveys further note that historical documents continue to be underrepresented in OCR evaluation and that existing benchmarks often fail to capture failure modes particularly relevant for archival collections~\cite{beyene2026}.

\noindent\textbf{OCR Evaluation Beyond Global Transcription Fidelity:}
OCR transcriptions are the input to numerous downstream applications, including information retrieval, named entity recognition, entity linking, and retrieval-augmented generation~\cite{greif2025multimodalllmsocrocr,sun2026goodocr}. Consequently, OCR errors propagate beyond transcription itself, and numerous studies have quantified their impact by evaluating complete OCR-to-application pipelines or by introducing synthetic OCR degradations with controlled error patterns~\cite{assessingOCRNLP,whenPostCorrectionNER,bazzo2020,hamdi2020assessing}.

These studies consistently show that the practical impact of OCR errors depends not only on their quantity but also on their nature and the information they affect. Beyond overall edit distance,~\cite{deepStatisticalOCR} analyzed OCR error distributions to better understand their implications for post-OCR correction, while other studies have argued that errors affecting named entities or retrieval terms have a disproportionate impact on downstream applications~\cite{beyondCERWER}. Recent analyses of VLM-based OCR have further shown that low edit distance may coexist with orthographic normalization, semantic substitutions, or hallucinated content that alter the information conveyed by historical documents~\cite{vesalainen2026,karamolegkou2026,gardella2026lowcerenoughanalysis}. Together, these observations suggest that OCR evaluation should consider not only how many errors are produced, but also whether the information most relevant to downstream use has been faithfully preserved.

In this work, we complement global transcription metrics with entity-level measures that analyze how OCR errors affect the fidelity and recoverability of information-bearing entities, independently of downstream extraction models.

\section{Dataset}

The Berrutti dataset~\cite{belzarena2025improving} was released in 2025 and is derived from the Berrutti Archive, a large microfilm collection containing over 3 million pages, mainly produced during the Uruguayan military dictatorship (1973--1985) by government agencies. The archive represents an important resource for historical research, human rights investigations, and transitional justice processes in Uruguay. Previous studies have emphasized the role of repressive archives in reconstructing state violence, identifying victims, and supporting truth and justice initiatives~\cite{
caetano2017,
lessa2022}. Therefore, accurate transcription of these documents is not only an OCR challenge but also a prerequisite for reliable historical analysis.


For research purposes, the Uruguayan \textit{Institución Nacional de Derechos Humanos y Defensoría del Pueblo} authorized the public release of a curated subset of 175 document images, which is the dataset used in this work. The selected documents were manually reviewed to ensure compliance with ethical and legal requirements and to provide a representative sample of the collection. 
Each document is provided with a full-text transcription, which enables document-level OCR evaluation.

In addition, we introduce entity-level annotations to evaluate OCR systems beyond global text similarity metrics. These annotations comprise five entity categories: 


\begin{itemize}[leftmargin=*]
    \item \texttt{PERSON}: names of individuals.
    \item \texttt{DATE/TIME}: explicit dates and time references.
    \item \texttt{LOCATION}: geographic references such as countries, cities, and other named places, as well as their demonyms.
    \item \texttt{ORGANIZATION}: explicitly named institutions, groups, newspapers, and other entities.
    \item \texttt{ID}: alphanumeric identifiers associated with people, organizations, documents, laws, or archival references.
\end{itemize}
%
%
We followed a conservative strategy aimed at maximizing consistency and minimizing ambiguity: generic references to institutions or social groups are excluded, while named entities are annotated using the shortest meaningful span. 

\begin{figure}[t!]
     \centering
\includegraphics[width=\linewidth]{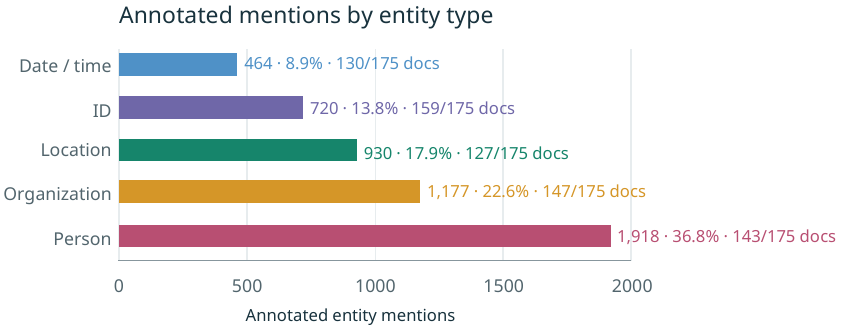}
    \vspace*{-0.6cm}
    \caption{Annotation statistics. Bars indicate the total number of annotated entity mentions for each category, while the accompanying values report the number of documents containing at least one instance of that entity type. 
    }
    \label{fig:entity-stats}
    \vspace*{-0.3cm}
\end{figure}

In total, the dataset contains 5,209 annotated entity mentions, corresponding to 4,383 distinct entities. The number of annotated entities per document varies substantially, ranging from 2 to 375, with a mean of 29.8 entity mentions per document. Fig.~\ref{fig:entity-stats} summarizes the distribution of the annotations by entity type. The dataset is publicly available at
\url{https://github.com/camilomarino/ocr_berrutti_dataset}.

\section{Experiments}

\noindent\textbf{Methods:} We evaluated a broad range of open-source systems. 
Traditional 
OCR systems include PP-OCRv5~\cite{ppocrv5}, EasyOCR~\cite{easyocr}, DocTR~\cite{doctr2021} and Tesseract~\cite{tesseract}. Specialized VLMs include recent VLMs designed specifically for document analysis and OCR, such as Surya-OCR~\cite{paruchuri2025surya}, Hunyuan-OCR~\cite{hunyuanvisionteam2025hunyuanocrtechnicalreport}, MinerU2.5-Pro~\cite{mineru25p}, Chandra-OCR 2~\cite{chandra}, GLM-OCR~\cite{duan2026glmocrtechnicalreport}, Qianfan-OCR~\cite{qianfan}, Dots-MOCR~\cite{dotsMOCR}, Dots-OCR~\cite{dotsOCR}, PaddleOCR-VL~\cite{paddleocrvl15multitask09bvlm}, and DeepSeek-OCR~\cite{wei2025deepseek}. Finally, general-purpose VLMs include Qwen 3.6~\cite{qwen3_6_alibaba}, Gemma 4~\cite{google_gemma4_docs}, and MiniCPM~\cite{yu2025minicpmv45}.

The evaluated systems span a wide range of OCR approaches, from traditional recognition pipelines to large multimodal generative models. Given the privacy constraints associated with the Berrutti archive, we limited our evaluation to models that can be run locally without transmitting document images to external services, even though the evaluated subset is publicly available under appropriate authorization.
Models with fixed OCR prompts were evaluated using their default instructions. Prompt-based models were evaluated using the same OCR-oriented prompt given below.

\begin{tcolorbox}[colback=gray!5, colframe=gray!50, title=\small{OCR-oriented Prompt},
    left=5mm,
    right=5mm,
    top=1mm,
    bottom=1mm,
    boxsep=1mm]
\linespread{0.8}
\footnotesize
OCR this scanned historical Uruguayan document page. It is a typewritten document from the Uruguayan dictatorship period.

Return only plain text in reading order. Preserve original line breaks, spelling, accents, capitalization, punctuation, abbreviations, dates, numbers, and hyphenation.

Read top-to-bottom. In multi-column or table-like regions, use the natural human order: columns left-to-right; rows top-to-bottom and cells left-to-right.

Do not explain. Do not summarize. Do not use markdown. Do not correct or expand abbreviations. Do not invent missing text. Use [ilegible] only for unreadable text.
\end{tcolorbox}

\noindent\textbf{Metrics:} We evaluated the transcriptions at two levels: full text and annotated entities. Before evaluation, we applied the same normalization to
the ground truth (GT) and the OCR prediction. This normalization removes
presentation-only Markdown and HTML markup, canonicalizes line endings,
collapses horizontal whitespace, trims individual lines, and collapses
consecutive blank lines. The remaining line structure, case, diacritics,
hyphens, and punctuation are preserved.


\paragraph*{Full-text Character Error Rate}

For each document $d$, we compute the minimum-distance Levenshtein edit script between the GT and OCR prediction. Let $S_d$, $D_d$, and $I_d$ denote the numbers of substitutions, deletions, and insertions, respectively, and let $N_d$ be the number of GT characters. We report the micro-averaged Character Error Rate (CER):
\begin{equation}
\mathrm{CER} =
\frac{\sum_d (S_d+D_d+I_d)}
{\sum_d N_d}.
\end{equation}

We report micro-averaged metrics because the number of annotated entity mentions varies substantially across documents. Micro-averaging weights each evaluated character or entity mention equally across the dataset, providing a direct measure of overall OCR performance. For completeness, document-wise macro-averaged results are available in~\cite{gardella2026lowcerenoughanalysis}.

\paragraph*{Entity Character Error Rate}
To measure errors affecting annotated entities, we compute Entity CER by
restricting the edit operations to characters belonging to annotated entities. Let $\mathcal{E}_d$ be the set of GT character positions covered by
entities. We count substitutions and deletions whose GT positions belong to
$\mathcal{E}_d$, and insertions occurring between two adjacent characters in
$\mathcal{E}_d$. The micro-averaged Entity CER is
\begin{equation}
\mathrm{Entity\,CER} =
\frac{\sum_d
(S_{d,\mathcal{E}}+D_{d,\mathcal{E}}+I_{d,\mathcal{E}})}
{\sum_d |\mathcal{E}_d|}.
\end{equation}

\paragraph*{Entity Coverage}
Entity CER measures errors assigned to entity regions by the global alignment,
but does not indicate whether an entity remains recoverable. We therefore complement it with entity Coverage@k. 
Let $M_d$ denote the number of annotated entity mentions in document $d$, and
let $M_{d,k}$ denote the number of mentions recovered with at most $k$
character edits. Dataset-level Coverage@k is defined as
\begin{equation}
\mathrm{Cov}@k =
\frac{\sum_d M_{d,k}}{\sum_d M_d}.
\end{equation}

A mention is considered \textit{recovered} when its normalized value matches a
contiguous, boundary-aligned substring of the normalized OCR transcription at
an edit distance no greater than $k$. Candidate substrings cannot begin
or end inside an alphanumeric sequence. Each distinct, non-overlapping OCR occurrence can recover at most one mention of the same normalized value. We evaluate $k\in\{0,\ldots,5\}$, where $\mathrm{Cov}@0$ corresponds to exact entity matching.

Unlike Entity CER, Coverage does not depend on the entity position induced by
the global alignment, but on whether the entity value remains recoverable from
the OCR transcription. 

\setlength{\tabcolsep}{3pt}
\begin{table}[t]
\centering
\begin{tabular}{lccccccc}
\toprule
\multirow{2}{*}{Model} &
\multirow{2}{*}{CER (\%)} &
\multicolumn{6}{c}{Entity CER (\%)} \\
\cmidrule(lr){3-8}
&
&
All &
Date &
ID &
Loc. &
Org. &
Person \\
\midrule
Surya-OCR & \textbf{8.15} & \textbf{9.91} & 12.67 & 26.67 & \underline{2.96} & \textbf{4.84} & \textbf{11.90}\\
Qwen 3.6 & \underline{9.78} & \underline{11.51} & \textbf{7.67} & \textbf{14.32} & \textbf{2.74} & \underline{4.98} & \underline{17.05}\\
PP-OCRv5 & 10.73 & 15.14 & 19.76 & 23.09 & 7.09 & 9.36 & 18.53\\
DocTR & 10.82 & 16.25 & 12.12 & \underline{14.66} & 6.00 & 6.93 & 24.08\\
MinerU2.5-Pro & 11.11 & 20.60 & 11.03 & 89.58 & 3.98 & 6.51 & 24.92\\
MiniCPM & 11.17 & 23.64 & 10.45 & 39.89 & 7.05 & 9.81 & 34.30\\
Dots-MOCR & 12.86 & 24.74 & 12.54 & 48.01 & 5.38 & 7.54 & 36.82\\
Gemma 4 & 12.99 & 29.75 & 12.06 & 48.56 & 6.92 & 8.66 & 45.77\\
PaddleOCR-VL & 13.85 & 25.72 & 16.71 & 41.54 & 6.93 & 11.21 & 36.80\\
Qwen 3.6$^\star$ & 14.08 & 28.05 & 17.82 & 45.43 & 18.71 & 15.70 & 35.73\\
GLM-OCR & 14.54 & 32.45 & 13.00 & 76.38 & 3.88 & 7.56 & 49.12\\
Chandra-OCR & 14.62 & 20.93 & \underline{8.56} & 30.54 & 3.46 & 5.90 & 33.06\\
Hunyuan-OCR & 15.02 & 33.57 & 10.12 & 94.78 & 4.32 & 6.29 & 50.10\\
Tesseract & 20.34 & 24.35 & 27.11 & 45.19 & 14.94 & 16.72 & 27.75\\
Dots-OCR & 21.65 & 33.85 & 16.61 & 96.53 & 5.24 & 8.38 & 48.41\\
DeepSeek-OCR & 23.93 & 45.97 & 15.19 & 132.21 & 6.54 & 10.10 & 67.26\\
Qianfan-OCR & 24.84 & 33.30 & 10.70 & 44.78 & 33.79 & 17.53 & 42.28\\
EasyOCR & 28.79 & 39.37 & 35.18 & 62.10 & 25.97 & 26.03 & 47.18\\
\bottomrule
\end{tabular}
\vspace{0.1cm}
\caption{CER and Entity CER (\%) by entity type on the Berrutti dataset. Best results are in bold and second-best results are underlined. The $\star$ in Qwen 3.6 indicates reasoning mode.}
\label{tab:berrutti-ocr-ranked}
\vspace*{-0.7cm}
\end{table}

\noindent\textbf{Results:}
Tab.~\ref{tab:berrutti-ocr-ranked} reports the CER and Entity CER values for the evaluated OCR systems. Three observations emerge. First, Entity CER is consistently higher than the global CER, suggesting that transcription errors tend to concentrate on information-bearing entities. Second, entity difficulty is not uniform: dates and locations are generally transcribed more accurately than person names and identifiers, independently of the OCR model. Finally, OCR rankings remain largely unchanged, with the systems achieving the lowest CER also obtaining the lowest Entity CER. Interestingly, traditional OCR engines remain competitive with VLM-based approaches in terms of CER and Entity CER.

Entity CER should nevertheless be interpreted together with layout-independent measures. Since it is computed from character-level alignments, it remains sensitive to reading-order and document layout errors. Complex layouts, such as tables or lists, may produce suboptimal alignments that increase the measured error even when the underlying entities remain recognizable. Consequently, higher Entity CER values should not always be interpreted as poorer entity transcription. To complement this view, we also evaluate entity Coverage, which measures whether relevant entities can still be recovered independently of alignment effects.
Tab.~\ref{tab:berrutti-cov-ranked} reports entity Coverage at zero edit distance, 
measuring the proportion of annotated entity mentions found exactly in the OCR output. Unlike Entity CER, Coverage substantially changes the ranking of OCR systems. VLM-based approaches generally achieve higher exact entity recovery, with Qwen 3.6 preserving more than 85\% of the annotated entities. Coverage also reveals differences across entity categories. Person names and identifiers are consistently the most challenging entities for VLM-based models, whereas dates and locations are recovered more reliably. Interestingly, Qwen 3.6 exhibits a considerably smaller degradation on these difficult categories than the remaining VLMs, suggesting greater robustness when contextual information is limited. In contrast, traditional OCR systems show comparatively higher identifier coverage despite lower overall performance, indicating that the two OCR paradigms produce qualitatively different error profiles.

\begin{table}[t]
\centering
\setlength{\tabcolsep}{5.5pt}
\begin{tabular}{lcccccc}
\toprule
\multirow{2}{*}{Model} &
\multicolumn{6}{c}{$\mathrm{Cov}@0$ (\%)} \\
\cmidrule(lr){2-7}
& All & Date & ID & Loc. & Org. & Person \\
\midrule
Qwen 3.6 & \textbf{85.39} & \textbf{86.64} & \textbf{79.17} & \underline{90.65} & \underline{85.39} & \textbf{84.88}\\
Chandra-OCR & \underline{73.09} & \underline{84.91} & 58.75 & \textbf{90.97} & \textbf{86.07} & 58.97\\
Qwen 3.6$^\star$ & 68.02 & 78.66 & 57.92 & 77.42 & 77.06 & \underline{59.12}\\
MinerU2.5-Pro & 67.54 & 78.02 & 54.72 & 89.14 & 80.97 & 51.09\\
Surya-OCR & 67.19 & 73.49 & 68.06 & 83.44 & 73.92 & 53.34\\
GLM-OCR & 66.29 & 80.82 & 27.78 & 89.46 & 81.65 & 56.57\\
MiniCPM & 65.56 & 79.31 & 57.78 & 82.90 & 75.96 & 50.36\\
Dots-MOCR & 65.52 & 74.57 & 43.61 & 87.20 & 82.33 & 50.73\\
Dots-OCR & 63.72 & 76.51 & 52.50 & 86.02 & 76.64 & 46.09\\
Qianfan-OCR & 63.66 & 78.88 & 61.94 & 83.44 & 76.55 & 43.12\\
Hunyuan-OCR & 63.58 & 81.68 & 45.28 & 85.05 & 80.12 & 45.52\\
Gemma 4 & 59.38 & 75.22 & 47.36 & 83.44 & 71.28 & 41.08\\
DocTR & 58.99 & 63.58 & \underline{70.14} & 64.52 & 55.73 & 53.02\\
PaddleOCR-VL & 57.94 & 70.47 & 21.81 & 83.23 & 70.60 & 48.44\\
DeepSeek-OCR & 55.65 & 71.55 & 28.19 & 80.43 & 71.88 & 40.15\\
Tesseract & 49.36 & 47.84 & 35.42 & 65.48 & 53.95 & 44.32\\
PP-OCRv5 & 48.28 & 42.89 & 63.06 & 67.53 & 50.72 & 33.21\\
EasyOCR & 31.62 & 32.97 & 33.47 & 42.04 & 37.30 & 22.05\\
\bottomrule
\end{tabular}
\vspace{0.1cm}
\caption{$\mathrm{Cov}@0$ (\%) by entity type on the Berrutti dataset. Best results are highlighted in bold and second-best results are underlined. The $\star$ in Qwen 3.6 indicates reasoning mode.}
\label{tab:berrutti-cov-ranked}
\vspace*{-0.7cm}
\end{table}



\begin{figure}[t!]
     \centering
\includegraphics[width=0.9\linewidth]{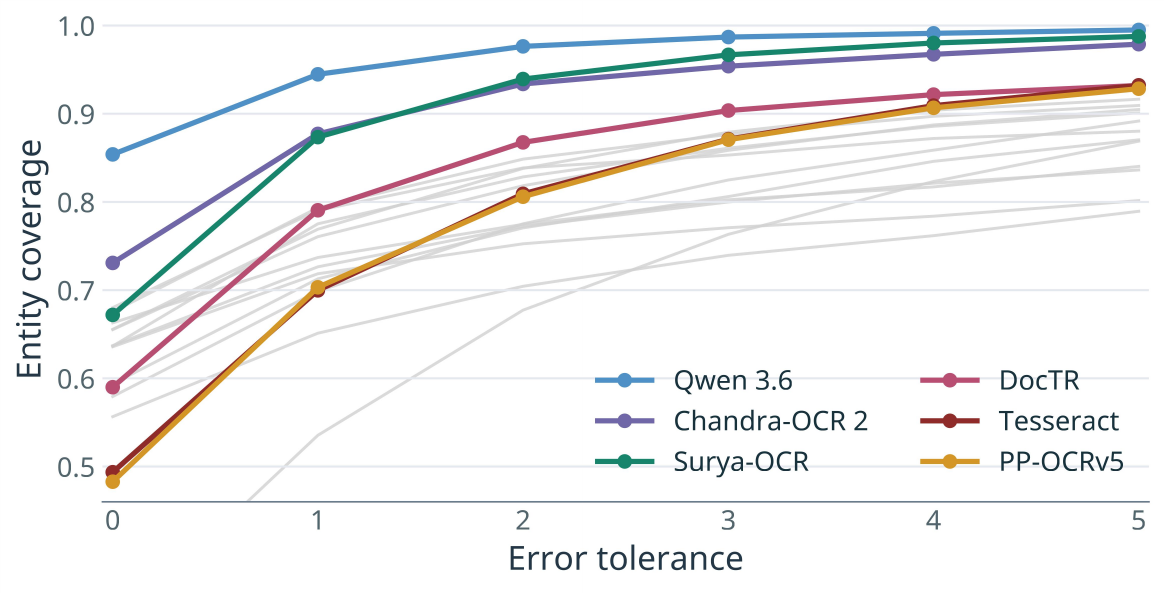}
\vspace*{-0.3cm}
\caption{Entity coverage as a function of the tolerated edit distance $k$. The highlighted curves correspond to representative OCR systems, while the gray curves show the remaining evaluated methods.}
\label{fig:cov}
\vspace*{-0.3cm}
\end{figure}

Fig.~\ref{fig:cov} extends the previous analysis by reporting entity Coverage as a function of the tolerated edit distance. As expected, coverage increases monotonically for all methods as larger transcription errors are accepted. Qwen 3.6 consistently achieves the highest coverage across all tolerance levels, indicating that it preserves the largest fraction of entities both exactly and under relaxed matching. However, the relative ordering of the remaining systems evolves with increasing tolerance, revealing differences in the severity of their transcription errors that are not captured by $\mathrm{Cov}@0$ alone.

Several models exhibit substantial gains after allowing only one or two character errors. At $k=2$, DocTR rises from thirteenth to fourth place, while Surya-OCR overtakes Chandra-OCR. These trends suggest that a considerable fraction of their entity mismatches correspond to localized transcription errors rather than completely incorrect entities. Manual inspection of DocTR outputs indicates that many of these differences are caused by Spanish-specific characters, particularly accented vowels and the letter \emph{ñ}, producing near-correct transcriptions that are recovered under relaxed matching.

These gains in entity preservation should also be considered in relation to computational cost. In our experiments, Qwen 3.6 required 53.9 s/page, compared with 0.57 s/page for DocTR (94× slower), highlighting a trade-off between transcription fidelity and scalability for large-scale archival digitization.



\section{Conclusions}

Our experiments show that entity preservation cannot be inferred solely from global transcription accuracy. While $\mathrm{Entity\,CER}$ largely follows the trends observed with CER, entity Coverage reveals substantial differences in the ability of OCR systems to preserve information exactly, leading to different rankings across methods. Together, these results indicate that transcription fidelity and information preservation represent related but distinct aspects of OCR quality.

Results also show systematic differences in the transcription of entity categories. Names and identifiers remain consistently more challenging than dates or locations across most evaluated systems, suggesting that the difficulty is driven not only by OCR quality but also by the intrinsic redundancy of the information being transcribed. Differences across entity categories should be interpreted in light of their length and redundancy: while longer entities such as person names or organizations remain identifiable despite localized errors, shorter identifiers can be severely affected by single-character errors. Furthermore, traditional OCR systems and VLM-based approaches exhibit different error characteristics, emphasizing that similar global accuracy may conceal substantially different behaviors.

A limitation of the proposed evaluation is that Coverage only measures the recovery of ground-truth entities and therefore primarily captures recall. It does not penalize hallucinated or spurious entities, which recent work, including our previous analysis~\cite{gardella2026lowcerenoughanalysis}, has shown to be an important failure mode of VLM-based OCR. Extending entity-aware evaluation to jointly quantify entity preservation and entity hallucination constitutes an important direction for future work. 

Moreover, the interpretation of Coverage depends on the matching criterion and the intrinsic properties of each entity category. A uniform character-level tolerance does not necessarily represent the same degree of preservation across different information types, as entities vary in structure, length, redundancy, and downstream relevance. Future work should investigate entity-aware evaluation criteria that jointly consider entity preservation, hallucination, and graded recoverability.





\section*{Acknowledgments}

The research that originated the results presented in this publication was partly supported by the Agencia Nacional de Investigación e Innovación (ANII) of Uruguay, the Programa de Desarrollo de Ciencias Básicas (PEDECIBA) of Uruguay, the France 2030 CollabNext project, and Université Paris-Saclay through the 2025 International Doctoral Cotutelle Funding Programme (ADI).

\bibliographystyle{IEEEtran}
\bibliography{refs}
\end{document}